\documentclass[letterpaper, 10 pt, conference]{ieeeconf}  

\IEEEoverridecommandlockouts                              

\makeatletter
\let\NAT@parse\undefined
\makeatother
\usepackage[numbers,sort&compress]{natbib}

\usepackage{times}
\usepackage{epsfig}
\usepackage{graphicx}
\usepackage{amsmath}
\usepackage{amssymb}
\usepackage[font=scriptsize]{caption}
\usepackage{subcaption}
\usepackage[ruled,vlined]{algorithm2e}
\usepackage{authblk}
\usepackage[T1]{fontenc}
\usepackage{xcolor}
\usepackage{balance}
\usepackage{calc}
\usepackage{lipsum}
\usepackage{hyperref}
\usepackage{adjustbox}

\title{\LARGE \bf
Probabilistic 3D Multi-Modal, Multi-Object Tracking\\ for Autonomous Driving
}

\author{Hsu-kuang Chiu$^1$, Jie Li$^2$, Rareș Ambruș$^2$ and Jeannette Bohg$^1$
\thanks{$^{1}$Stanford University, %
$^2$Toyota Research Institute
}
\thanks{Toyota Research Institute ("TRI") provided funds to assist the authors with their research but this article solely reflects the opinions and conclusions of its authors and not TRI or any other Toyota entity.}
}

\begin{document}
\balance

\maketitle
\thispagestyle{empty}
\pagestyle{empty}

\begin{abstract} 
Multi-object tracking is an important ability for an autonomous vehicle to safely navigate a traffic scene. Current state-of-the-art follows the tracking-by-detection paradigm where existing tracks are associated with detected objects through some distance metric. Key challenges to increase tracking accuracy lie in data association and track life cycle management. We propose a probabilistic, multi-modal, multi-object tracking system consisting of different trainable modules to provide robust and data-driven tracking results. First, we learn how to fuse features from 2D images and 3D LiDAR point clouds to capture the appearance and geometric information of an object. Second, we propose to learn a metric that combines the Mahalanobis and feature distances when comparing a track and a new detection in data association. And third, we propose to learn when to initialize a track from an unmatched object detection. Through extensive quantitative and qualitative results, we show that when using the same object detectors our method outperforms state-of-the-art approaches on the NuScenes and KITTI datasets.

\end{abstract}

\section{Introduction}
3D multi-object tracking is essential for autonomous driving. It estimates the location, orientation, and scale of all the traffic participants over time. By taking temporal information into account, a tracking module can filter outliers from frame-based object detection and be more robust to partial or full occlusions. 
The resulting trajectories may then be used to infer motion patterns and driving behaviours of each traffic participant to improve motion forecasting. This enables safe decision-making in autonomous driving.

Current state-of-the-art in 3D multi-object tracking~\cite{yin2020center, chiu2020probabilistic} follows the tracking-by-detection paradigm. These methods first use a 3D object detector to estimate the bounding box location and orientation of each object in each frame. Then they either use the center or Mahalanobis distance~\cite{mahalanobis1936distance} as data association metric between detections and existing tracks.
However, these metrics only evaluate the distance of objects and the differences of the bounding box size and orientation, while ignoring each object's geometric and appearance features. Therefore, data association performance highly depends on the accuracy of motion prediction. For objects that are hard to predict precisely, e.g. pedestrians, motorcycles or cars making sharp turns, the Euclidean distance between prediction and the correct detection can be high. Therefore, they may not be matched correctly. 
%
%
\cite{liang2020pnpnet, weng2020gnn3dmot} attempt to improve data association by learning an association metric from the tracker trajectories and the detection features. However, these methods are still unable to outperform the aforementioned simple method based on center distances~\cite{yin2020center}. The results indicate that building a neural network for effective data association is challenging. 


We propose to learn how to weigh the Mahalanobis distance~\cite{mahalanobis1936distance} and the distance based on geometric and appearance features when comparing a track and detection for data association. These features are extracted from 3D LiDAR point clouds and 2D camera images. Different from~\cite{liang2020pnpnet, weng2020gnn3dmot}, we use the learned metric within a standard Kalman Filter~\cite{kalman1960filter} that is effective for multi-object tracking~\cite{chiu2020probabilistic}. Additionally, a Kalman filter offers interpretability and explicit uncertainty estimates that can be used for down-stream decision-making. 

In addition to data association, track life cycle management is another important component of online tracking systems. Track life cycle management determines when to initialize and terminate each track. This decision significantly affects the number of false positives and identity switches. However, track life cycle management has not attracted much attention from the research community. Prior works either initialize a new track for every unmatched detection~\cite{yin2020center}, or create temporary tracks and convert them into full tracks given enough consecutive matches~\cite{weng2019ab3dmot, chiu2020probabilistic, shenoi2020jrmot, weng2020gnn3dmot, liang2020pnpnet}.


We propose to learn whether to initialize a new track from an unmatched detection based on its geometric and appearance features. This approach helps our tracking method avoid initializing new tracks for potential false-positives.

In summary, we propose a probabilistic, multi-modal, multi-object tracking system consisting of three trainable modules (\textbf{Distance Combination}, \textbf{Track Initialization} and \textbf{Feature Fusion}) to provide robust and data-driven tracking results. We evaluate our approach on the NuScenes~\cite{caesar2019nuscenes} and KITTI~\cite{geiger2012kitti} datasets using leading object detectors~\cite{yin2020center, shi2019pointrcnn} that take 3D LiDAR point clouds as input. We show that the proposed approach outperforms the tracking methods reported in~\cite{yin2020center} and~\cite{weng2020gnn3dmot}. By effectively fusing 2D and 3D input, we can increase performance gains even further. Our qualitative results also reveal significantly fewer false positive tracks which is important for decision-making. We expect further performance gains if using even newer object detectors as our approach is agnostic to this choice.

\section{Related Work}

\subsection{3D Object Detection}
Most 3D multi-object tracking systems~\cite{yin2020center, chiu2020probabilistic, weng2019ab3dmot, liang2020pnpnet, shenoi2020jrmot, weng2020gnn3dmot, zhou2020tracking} perform tracking on the detection bounding boxes provided by 3D object detectors. Therefore, the choice of 3D object detector is important for the overall performance of each tracking system. 3D object detection can be applied to camera images~\cite{chen2016monocular, brazil2019m3d}, LiDAR point clouds~\cite{  zhu2019megvii, zhou2018voxelnet, yan2018second, yan2018pixor, lang2019pointpillar, shi2019pointrcnn}, or their combination~\cite{liang2019multi, vora2020pointpainting, qi2018frustum}.
Monocular 3D object detection models are unlikely to be on par with models that utilize LiDAR or depth information. Therefore, 3D multi-object tracking algorithms~\cite{zhou2020tracking, hu2019joint} relying on monocular 3D object detectors are usually unable to outperform tracking methods relying on LiDAR- or depth-based object detectors.  

In our proposed tracking system, we use the CenterPoint 3D object detector~\cite{yin2020center} that is one of the top performers in the NuScenes Detection Challenge~\cite{caesar2019nuscenes}. Note that our method is agnostic to the detector. CenterPoint quantizes LiDAR point clouds and generates the feature map using PointNet~\cite{qi2019pointnet, qi2017pointnetplusplus}. The feature map is then fed to a key point detector for locating centers of objects and regressing the size and orientation of the bounding boxes.

\subsection{3D Multi-Object Tracking}
Most 3D multi-object tracking algorithms adopt the tracking-by-detection framework. They take 3D object detection results as input to the tracking methods. In the data association step, different distance metrics are used to find the matched track-detection pairs. 
For example, AB3DMOT~\cite{weng2019ab3dmot} uses the 3D intersection-over-union (3D IOU) as an extension to the 2D IOU in 2D tracking algorithms~\cite{bewley2016simple}.  
ProbabilisticTracking~\cite{chiu2020probabilistic} uses the Mahalanobis distance that takes uncertainty of the tracked state into account.
CenterPoint~\cite{yin2020center} uses the object center distance and achieves competitive tracking performance mainly due to a newly proposed 3D object detector that is better than the one used in \cite{weng2019ab3dmot,chiu2020probabilistic}. CenterPoint~\cite{yin2020center} is currently one of the leading methods in the NuScenes Tracking Challenge~\cite{caesar2019nuscenes}.

Several other 3D tracking methods proposed to combine the tracker trajectory with object geometric and appearance features. GNN3DMOT~\cite{weng2020gnn3dmot} uses a graph neural network and 2D-3D multi-feature learning for data association. PnPNet~\cite{liang2020pnpnet} presents an end-to-end trainable model to jointly solve detection, tracking, and prediction tasks. However, they are unable to outperform the aforementioned much simpler CenterPoint~\cite{yin2020center} algorithm on the NuScenes~\cite{caesar2019nuscenes} dataset.

Different from related work, we propose a probabilistic multi-modal multi-object tracking system that learns to combine the Mahalanobis distance~\cite{mahalanobis1936distance} and the deep feature distance for data association. We also propose a data-driven approach for track life cycle management. 

\section{Method}
A flowchart of our method is shown in Fig.~\ref{fig:arch_deep}. Building upon  ProbabilisticTracking~\cite{chiu2020probabilistic} as the baseline, our algorithm takes both LiDAR point clouds and camera images as input and conducts object tracking through Kalman Filters. Our proposed tracking algorithm features three trainable components to robustify data association and track life cycle management: 
the \textbf{Feature Fusion Module} merges the LiDAR and image features to generate the fused deep features. The \textbf{Distance Combination Module} learns to combine the deep feature distance with the Mahalanobis distance as the final metric for data association. Additionally, we also introduce the \textbf{Track Initialization Module} that learns to decide whether to initialize a new track for each unmatched detection based on the fused 2D and 3D deep features. In the following sections, we will describe each of the core components of our proposed tracking model.

\newdimen\figrasterwd
\figrasterwd\textwidth

\begin{figure*}[ht]
  \centering
  \parbox{\figrasterwd}{
    \parbox{.55\figrasterwd}{%
      \subcaptionbox{}{
      \centering
      \includegraphics[width=\hsize]{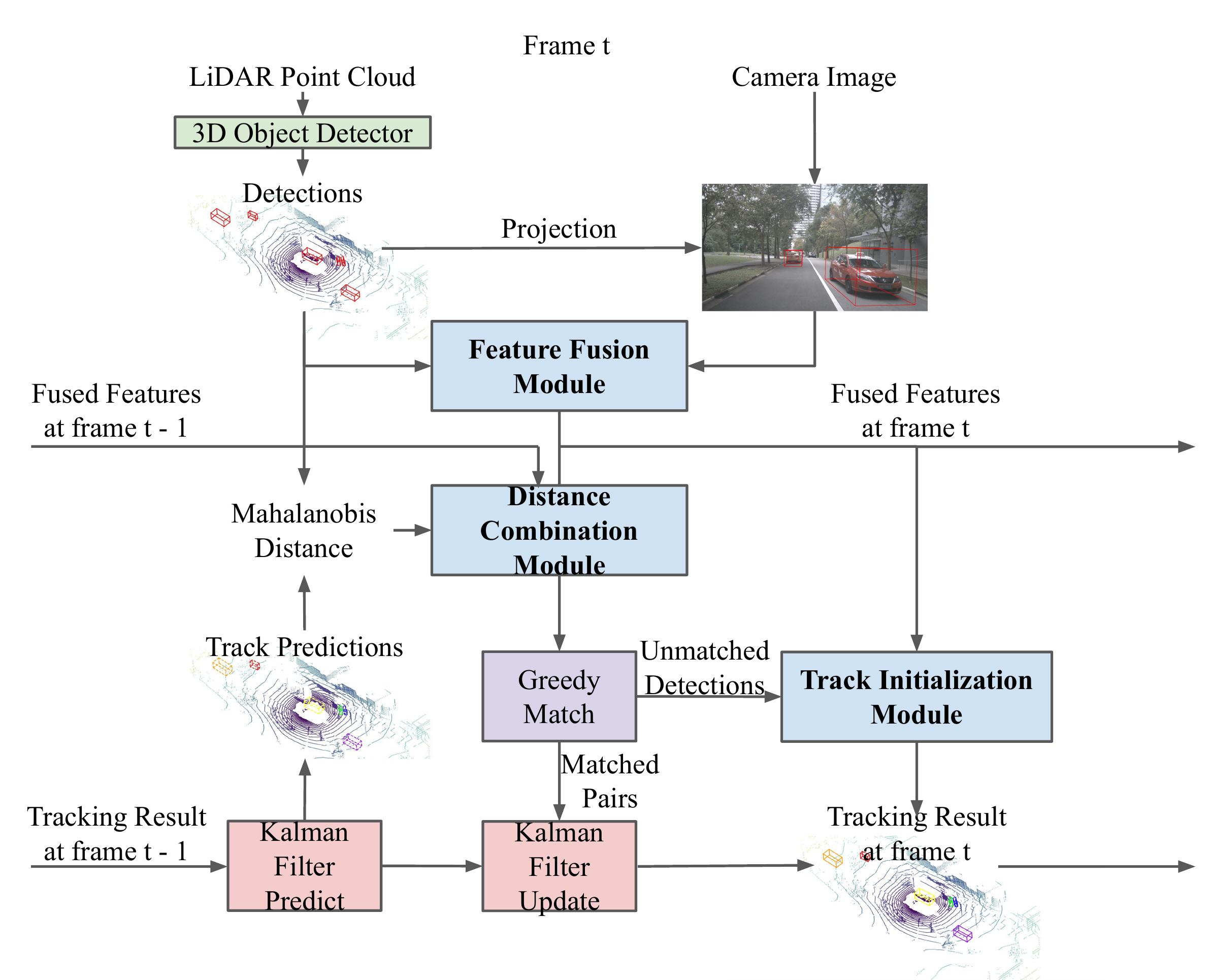}}
    }
    \hspace{5em}
    \parbox{.32\figrasterwd}{%
      \subcaptionbox{\label{fig:arch_1}}{\includegraphics[width=\hsize]{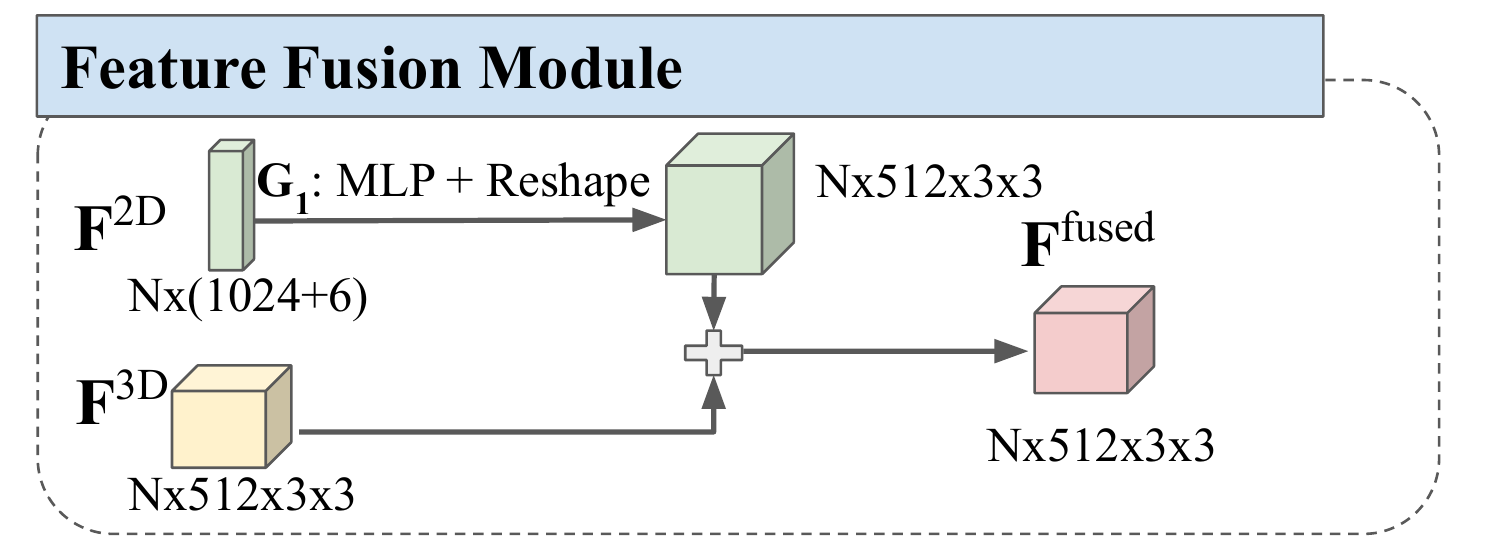}}
      \subcaptionbox{\label{fig:arch_2}}{\includegraphics[width=\hsize]{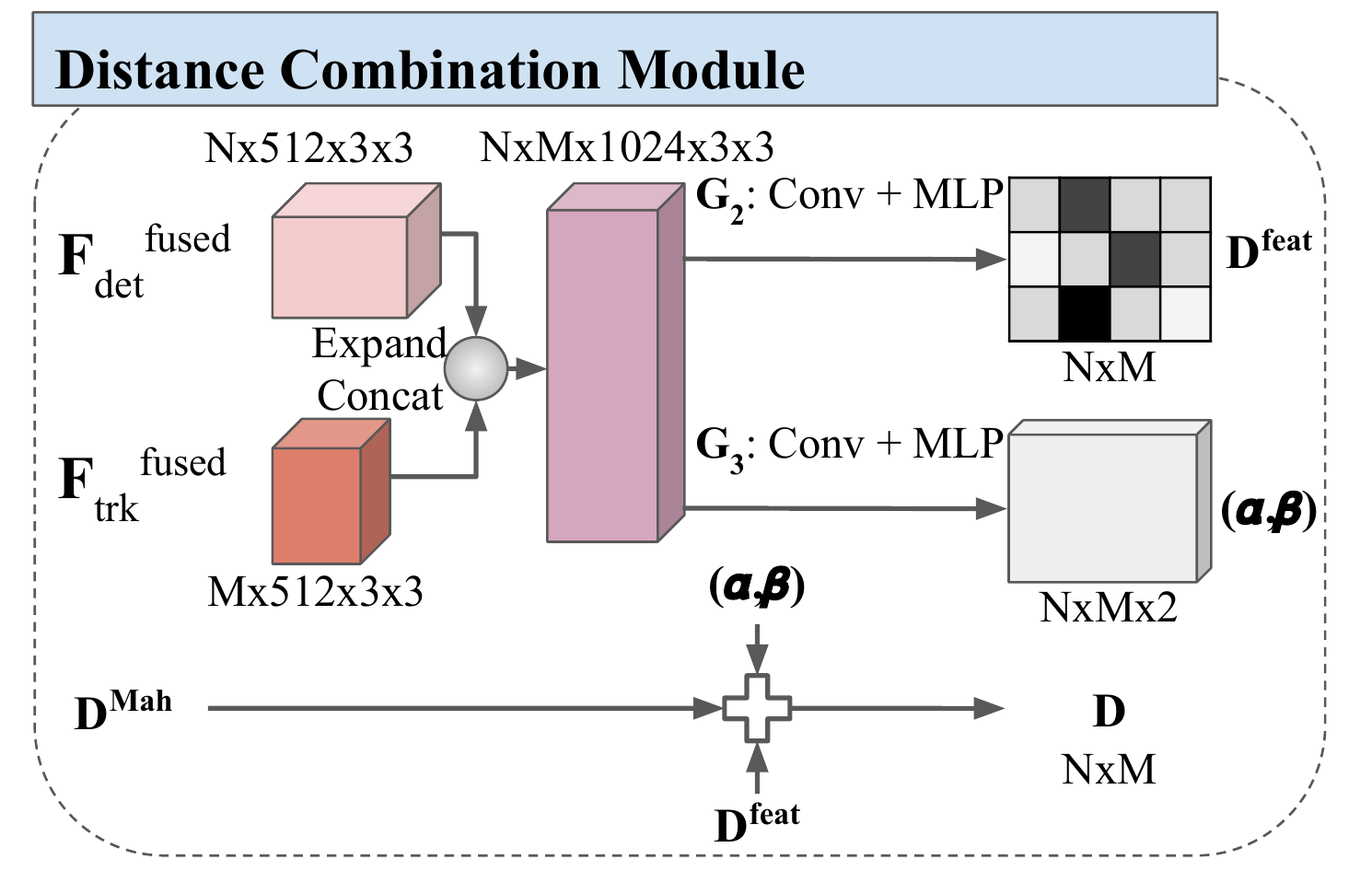}}
      \subcaptionbox{\label{fig:arch_3}}{\includegraphics[width=\hsize]{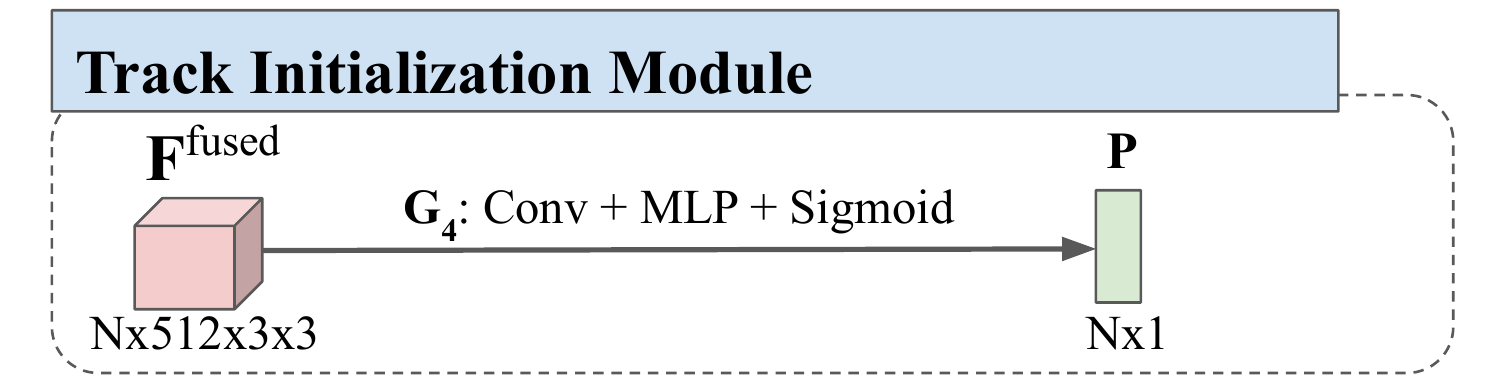}}
    }
  }
  \caption{Algorithm Flowchart. Sub-figure (a) depicts the high level overview of our proposed architecture, and (b)(c)(d) on the right illustrate the details of each neural network module. At frame $t$, We use a 3D object detector and extract the LiDAR and image features for each detected object. These features are fused by the \textbf{Feature Fusion Module}. The fused features of detections from time $t$ and tracks from time $t-1$ are used in the \textbf{Trainable Distance Combination Module} to learn the combination of the deep feature distance and the Mahalanobis distance. We apply a greedy match algorithm on the combined distance for data association. Matched pairs are further processed by the Kalman Filter to refine the final object states. The \textbf{Track Initialization Module} determines whether to initialize a new track for each unmatched detection.\vspace{-10pt}}
\label{fig:arch_deep}
\end{figure*}

\subsection{Kalman Filters}
We build upon prior work on ProbabilisticTracking~\cite{chiu2020probabilistic} and use Kalman Filters~\cite{kalman1960filter} for object state estimation. Each object's state is represented by 11 variables:
\begin{equation}\label{eq:state}
{\bf s}_t = (x, y, z, a, l, w, h, d_x, d_y, d_z, d_a)^T,
\end{equation}
where $(x, y, z)$ is the center position of the object's 3D bounding box, $a$ is the angle between the object's facing direction and the x-axis, $(l, w, h)$ represent the length, width, and height of the bounding box, and $(d_x, d_y, d_z, d_a)$ represent the difference of $(x, y, z, a)$ between the current and the previous frame.

We model the dynamics of the moving objects using a linear motion model, and assume constant linear and angular velocity as well as constant object dimensions, i.e. they do not change during the prediction step. Following the standard Kalman Filter formulation, we define the prediction step as:
\begin{align}
\label{eq:predicted_mean}
    {\bf \hat{\mu}}_{t+1} &= {\bf A} \mathbf{\mu}_t\\
    \label{eq:predicted_covariance}
    \hat{\Sigma}_{t+1} & = {\bf A} \Sigma_t {\bf A}^T + {\bf Q}
\end{align}
where ${\bf \mu}_t$ is the estimated mean of the true state $s$ at time $t$, and $\hat{\mu}_{t+1}$ is the predicted state mean at time $t+1$. The matrix ${\bf A}$ is the state transition matrix of the process model. The matrix $\Sigma_t$ is the state covariance at time $t$, and $\hat{\Sigma}_{t+1}$ is the predicted state covariance at time $t+1$. The matrix ${\bf Q}$ is the process model noise covariance.

We use CenterPoint~\cite{yin2020center}'s 3D object detector to provide the observations to our Kalman Filter. The per frame 3D object detection results consist of a set of 3D bounding boxes, with each box being represented by 9 variables: 
\begin{equation}\label{eq:observation}
{\bf o}_t = (x, y, z, a, l, w, h, d_x, d_y)^T,
\end{equation}
where $(x, y, z, a, l, w, h)$ are the bounding box's center position, orientation, and scale, similar to the definitions in Equation \ref{eq:state}. The remaining two variables $(d_x, d_y)$ represent the difference of $(x, y)$ between the current frame and the previous frame. These two values can be derived by multiplying the detector's estimated center velocity with the time duration between two consecutive frames. 
We use a linear observation model ${\bf H}$ with additive Gaussian noise that has zero mean and noise covariance ${\bf R}$. Using this observation model and the predicted object state ${\bf \hat{\mu}}_{t+1}$, we can predict the next measurement ${\bf \hat{o}}_{t+1}$ and innovation covariance ${\bf S}_{t+1}$ that represents the uncertainty of the predicted object detection:
\begin{align}
    {\bf \hat{o}}_{t+1} & = {\bf H} {\bf \hat{\mu}}_{t+1}\\
    {\bf S}_{t+1} & = {\bf H} \hat{\Sigma}_{t+1} {\bf H}^T + {\bf R} \label{eq:S}
\end{align}
The noise covariance matrices {\bf Q} and {\bf R} of the process model and the observation model are estimated from the statistics of the training set data, as proposed in~\cite{chiu2020probabilistic}.

\subsection{Fusion of 2D and 3D features}

This module is designed to fuse the features from 2D camera images and 3D LiDAR point clouds per detection in the key frames. The fused features will be used as input to the \textbf{Distance Combination Module} and the \textbf{Track Initialization Module}. 
For each detection, we first map its 2D position $(x, y)$ from the world coordinate system to the 2D location $(x_{map}, y_{map})$ in the 3D object detector's intermediate feature map coordinate system. From this intermediate feature map, we extract a $512\times3\times3$ LiDAR point cloud feature. Instead of only extracting a single feature vector located at $(x_{map}, y_{map})$ in the feature map, we extract all the feature vectors inside the associated $(3 \times 3)$ $xy$ region centered at $(x_{map}, y_{map})$ in order to utilize more context information.

We then project the 3D detection bounding box to the camera image plane, and extract the corresponding 2D image feature from a COCO~\cite{lin2014microsoft} pre-trained Mask R-CNN~\cite{he2017mask}. For each projected 2D bounding box, we extract a 2D image feature that concatenates a $1024$ dimensional vector from the RoIAlign feature of the projected 2D bounding box and a $6D$ one-hot vector indicating to which camera plane (out of 6 in the sensor sweep) the object projects.

Finally, we combine the two feature vectors through a multi-layer-perceptron (MLP) and a reshape operation:
\begin{equation}\label{eq:fusion_detection}
\textbf{F}^{fused} = \mathbf{G}_1(\textbf{F}^{2D}) + \textbf{F}^{3D}, 
\end{equation}
where $\textbf{F}^{fused}\in \mathbb{R}^{N\times512\times3\times3}$ is the fused feature of $N$ detections; $\textbf{F}^{2D} \in \mathbb{R}^{N\times(1024+6)}$ is the 2D feature; $\textbf{F}^{3D} \in \mathbb{R}^{N\times512\times3\times3}$ is the 3D feature; ${\bf G}_1(\cdot)$ denotes the MLP and the reshape operation depicted in Figure~\ref{fig:arch_1}. This MLP has a hidden size $1536$
and an output size $4608$
, using a Rectified Linear Unit (ReLU) as the activation function. Note that we do not train this \textbf{Feature Fusion Module} by itself. Instead, we connect it to the \textbf{Distance Combination Module} and \textbf{Track Initialization Module} and train with the two modules.

\subsection{Distance Combination Module}
This module provides a learned distance metric for data association between N detections and M tracks. The metric combines information from state estimates as well as appearance and geometry features. Specifically, we design a linear combination of the  Mahalanobis and deep feature distance:
\begin{equation}\label{eq:combine}
\textbf{D} = \textbf{D}^{Mah} + \boldsymbol\alpha \odot (\textbf{D}^{feat} - (0.5 + {\boldsymbol\beta})),
\end{equation}
where $\textbf{D}^{Mah}\in \mathbb{R}^{N\times M}$ denotes the Mahalanobis distance matrix where each element contains the distance between each detection and predicted state per track;  $\textbf{D}^{feat}\in \mathbb{R}^{N\times M}$ denotes the feature distance matrix whose elements measure the feature dissimilarity between each detection and each track, and $(\boldsymbol\alpha,\boldsymbol\beta )$, each with shape $\mathbb{R}^{N\times M}$, are the combining coefficient matrices. The symbol $\odot$ denotes the element-wise product operator. The constant $0.5$ serves as the initial bias term of the linear combination to help the model training converge faster. Each element of $\textbf{D}^{Mah}$ is computed by:
\begin{equation}
\textbf{D}^{Mah}_{(n,m)} = \sqrt{({\bf o_n} - {\bf H}\hat{\mu}_m)^{T} {{\bf S}}^{-1} ({\bf o_n} - {\bf H}\hat{\mu}_m)},
\end{equation}
where $\bf o_{n}$ is the detection $n$, defined in Eq.~\ref{eq:observation}, $\textbf{H}$ is the linear observation model, $\hat{\mu}_m$ is the $m$th track predicted state mean, and $\textbf{S}$ is the innovation covariance matrix as defined in Eq.~\ref{eq:S}.

We employ a two-stage training approach on a neural network, as depicted in Figure~\ref{fig:arch_2}, to first learn the deep feature distance $\textbf{D}^{feat}$ and then the coefficient matrices $(\boldsymbol\alpha,\boldsymbol\beta )$ to generate the final combined distance metric $\textbf{D}$.

\subsubsection{Deep Feature Distance}
The network learns an $N\times M$ distance map from fused features of the $N$ detections and $M$ tracks $(\textbf{F}_{det}^{fused}, \textbf{F}_{trk}^{fused})$ : 
\begin{equation}
    \textbf{D}^{feat.} = \mathbf{G}_2(\textbf{F}_{det}^{fused}, \textbf{F}_{trk}^{fused}),
\end{equation}
where $\mathbf{G}_2(\cdot)$, depicted in Figure~\ref{fig:arch_2}, denotes a convolutional layer with a kernal size $3\times3$ and an output channel size $256$, followed by a ReLU and an MLP layer with a hidden size $128$.
We supervise the feature distance learning by treating it as a  binary classification problem.
And we train the network with Binary Cross Entropy Loss:
\begin{equation}
    L^{dist} = \mathbf{BCE}(\textbf{D}^{feat}, \textbf{K}),
\end{equation}
where \textbf{K} is the supervising matching indicator matrix in which $0$ indicates a matched track-detection feature pair and $1$ indicates an unmatched feature pair.
Since there is no ground-truth annotation for each track-detection pair, we treat a pair as matched if the tracking box's closest ground-truth box in the previous frame and the detection box's closest ground-truth box in the current frame belong to the same object identity, and their 2D center Euclidean distances to their closest ground-truth boxes are all less than 2 meters.



\subsubsection{Combining Coefficients}

We fix the learned feature distance  $\textbf{D}^{feat}$ and then train the remaining part of the \textbf{Distance Combination Module} to learn the coefficient matrices $\boldsymbol\alpha$ and $\boldsymbol\beta$ so that they can adjust the final distance $\textbf{D}$ based on how important each deep feature distance is.
\begin{equation}
    (\boldsymbol\alpha,\boldsymbol\beta) = \mathbf{G}_3(\textbf{F}_{det}^{fused}, \textbf{F}_{trk}^{fused}),
\end{equation}
where $\mathbf{G}_3(\cdot)$ denotes the convolutional and MLP layers in Figure~\ref{fig:arch_2}. $\mathbf{G}_3(\cdot)$ has the similar network architecture as $\mathbf{G}_2(\cdot)$ except for the output channel size.
Inspired by PnPNet~\cite{liang2020pnpnet}, we train this module with a combination of max-margin and contrastive losses. For a pair of a positive sample $i$ and a negative sample $j$, we define its max-margin loss as follows:
\begin{equation}\label{eq:contrastive_max_margin}
L_{i,j}^{contr} = \mathbf{max}(0, C^{contr} - (d_i - d_j)),
\end{equation}
where $C^{contr}$ is a constant margin, $d_i$ is the combined distance of positive sample $i$ and $d_j$ is the combined distance of negative sample $j$ as can be found in distance matrix $D$ in Equation~\ref{eq:combine}.
The overall contrastive loss is given as follows:
\begin{equation}\label{eq:contrastive}
L^{contr} = \frac{1}{|\text{Pos}||\text{Neg}|} \sum_{i \in \text{Pos}, j \in \text{Neg}} L_{i,j}^{contr},
\end{equation}
where $\text{Pos}$ denotes the set of positive track-detection pairs and $\text{Neg}$ denotes the set of negative track-detection pairs. This loss function design encourages the neural network to learn to generate a distance $d_i$ for every positive track-detection sample to be smaller than the distance $d_j$ of any negative sample by adjusting the elements of $\alpha$ and $\beta$.

To also use the learned combined distance $D$ to reject unmatched outliers at inference time, we define two other max-margin losses for the positive sample set and negative sample set as follows:
\begin{equation}\label{eq:max_margin_pos}
L^{pos} = \frac{1}{|\text{Pos}|} \sum_{i \in \text {Pos}} \mathbf{max}(0, C^{pos} - (T - d_i)),
\end{equation}
\begin{equation}\label{eq:max_margin_neg}
L^{neg} = \frac{1}{\text{|Neg|}} \sum_{j \in \text{Neg}} \mathbf{max}(0, C^{neg} - (d_j - T)),
\end{equation}
where $C^{pos}$ and $C^{neg}$ denote constant margins and $T$ is the constant threshold used to reject unmatched outliers at inference time. This loss function design encourages the neural network to generate a distance $d_i$ smaller than the threshold $T$ for any positive sample, and a distance $d_j$ larger than $T$ for any negative sample.

The overall training loss of this neural network is defined as follows:
\begin{equation}\label{eq:loss_trainable_distance_combination_module}
L^{coef} = L^{contr} + L^{pos} + L^{neg}.
\end{equation}

In our implementation, we choose $T = 11$, the same threshold value used in~\cite{chiu2020probabilistic}. We set $C^{contr} = 6$, roughly half of $T$. And we set $C^{pos} = C^{neg} = 3$, half of $C^{contr}$.

At test time, once we calculate the combined distance, we conduct data association using the greedy matching algorithm from ProbabilisticTracking~\cite{chiu2020probabilistic}.

\subsection{Track Initialization Module}
Track life cycle management is another important component of multi-object tracking systems. Most prior works either always initialize a new track for every unmatched detection~\cite{yin2020center}, or create a temporary track and then wait for a constant number of consecutive matches before converting the temporary track to a full track~\cite{chiu2020probabilistic,weng2019ab3dmot, shenoi2020jrmot}. 

Different from the prior heuristic approaches, we treat the track initialization task as a binary classification problem. We propose the \textbf{Track Initialization Module}, which takes the fused feature $F^{fused}$ of unmatched detections as input, and generates an output confidence score $\textbf{P}$ on whether we should initialize a new track or not:
\begin{equation}
\textbf{P} = \mathbf{G}_4(\textbf{F}^{fused}),
\end{equation}
where $\mathbf{G}_4$ denotes the convolutional, MLP, and Sigmoid layers depicted in Figure~\ref{fig:arch_3}. Its couvolutional and MLP layers have the same architecture as $\mathbf{G}_2$.
We train $\mathbf{G}_4$ as a binary classifier using the Cross-Entropy loss:

\begin{equation}\label{eq:loss_feature}
L^{init} = \mathbf{BCE}(\textbf{P}, \textbf{P}^{target}),
\end{equation}
where $P_n^{target}=1$ if there is a ground-truth object close to detection $n$, otherwise $P_n^{target}=0$.
At inference time, we initialize an unmatched detection with new tracker if $P_n$ is larger than $0.5$. 
This \textbf{Track Initialization Module} helps our proposed tracking system reduce the number of false-positive tracks as depicted in Figure~\ref{fig:visual_motorcycle}.



\section{Experimental Results}
\label{sec:experimental_results}


\begin{table*}[ht!]
\small
\caption{Evaluation results on the NuScenes~\cite{caesar2019nuscenes} validation set: evaluation in terms of overall AMOTA and individual AMOTA for each object category, in comparison with the baseline methods. In each column, the best-obtained results are typeset in boldface. (*Our implementation by using~\cite{chiu2020probabilistic}'s open-source code and~\cite{yin2020center}'s object detection results.)
\vspace{-10pt}}
\label{tab:val_evaluation_results}
\begin{center}
\begin{adjustbox}{width=1\textwidth}
\begin{tabular}{ l|l|l|cccccccc}
  \hline
  Tracking method & Modalities & Input detection & Overall & bicycle & bus & car & motorcycle & pedestrian & trailer & truck \\
  \hline
  \hline
  AB3DMOT~\cite{weng2019ab3dmot} & 3D & MEGVII~\cite{zhu2019megvii} & 17.9 & 0.9 & 48.9 & 36.0 & 5.1 & 9.1 & 11.1 & 14.2 \\
  ProbabilisticTracking~\cite{chiu2020probabilistic} & 3D & MEGVII~\cite{zhu2019megvii} & 56.1 & 27.2 & 74.1 & 73.5 & 50.6 & 75.5 & 33.7 & 58.0\\
  \hline
  CenterPoint~\cite{yin2020center} & 3D & CenterPoint~\cite{yin2020center} & 65.9	& 43.7 & 80.2 & 84.2 & 59.2 & \textbf{77.3} & 51.5 & \textbf{65.4} \\
  ProbabilisticTracking~\cite{chiu2020probabilistic}* & 3D & CenterPoint~\cite{yin2020center} & 61.4 & 38.7 & 79.1 & 78.0 & 52.8 & 69.8 & 49.4 & 62.2 \\
  \hline
  Our proposed method & 3D & CenterPoint~\cite{yin2020center} & 67.7 & 47.0 & 81.9 & 84.2 & 66.8 & 75.2 & \textbf{53.5} & \textbf{65.4} \\
  \hline
  Our proposed method & 2D + 3D & CenterPoint~\cite{yin2020center} & \textbf{68.7} & \textbf{49.0} & \textbf{82.0} & \textbf{84.3} & \textbf{70.2} & 76.6 & 53.4 & \textbf{65.4} \\
  \hline
\end{tabular}
\end{adjustbox}
\end{center}
\vspace{-10pt}
\end{table*}

\begin{table}[ht!]
\small
\caption{Evaluation results on the NuScenes~\cite{caesar2019nuscenes} validation set: evaluation in terms of overall AMOTA and the AMOTA for car.
GNN3DMOT~\cite{weng2020gnn3dmot} only reports the overall AMOTA, and PnPNet~\cite{liang2020pnpnet} only reports the AMOTA for cars. Note that each method uses a different 3D object detector, and that could affect the tracking accuracy significantly. (*GNN3DMOT~\cite{weng2020gnn3dmot} renames AMOTA as sAMOTA in their paper.)
\vspace{-10pt}}
\label{tab:val_evaluation_partial}
\begin{center}
\begin{tabular}{ l|l|l|cccccccc}
  \hline
  Tracking method & Modalities & Overall & car \\
  \hline
  \hline
  GNN3DMOT~\cite{weng2020gnn3dmot}* & 2D + 3D & 29.84 & - \\
  PnPNet~\cite{liang2020pnpnet} & 2D + 3D & - & 81.5 \\ 
  Our proposed method & 2D + 3D & \textbf{68.7} & \textbf{84.3} \\
  \hline
\end{tabular}
\end{center}
\end{table}

\begin{table}[ht!]
\small
\caption{Evaluation results on the KITTI~\cite{geiger2012kitti} validation set: evaluation in terms of the AMOTA and MOTA for cars.
We follow ~\cite{weng2020gnn3dmot} by using the Point R-CNN~\cite{shi2019pointrcnn} 3D object detector and the same training-validation data split setting. (*GNN3DMOT~\cite{weng2020gnn3dmot} renames AMOTA as sAMOTA in their paper.)
\vspace{-10pt}}
\label{tab:val_evaluation_kitti}
\begin{center}
\begin{tabular}{ l|l|c|c}
  \hline
  Tracking method & Modalities & AMOTA & MOTA\\
  \hline
  \hline
  GNN3DMOT~\cite{weng2020gnn3dmot}* & 2D + 3D & 93.68 & 84.70 \\
  Our proposed method & 2D + 3D & \textbf{96.99} & \textbf{93.89} \\
  \hline
\end{tabular}
\end{center}
\end{table}

\subsection{Dataset}
We evaluate our method on the NuScenes~\cite{caesar2019nuscenes} and KITTI~\cite{geiger2012kitti} datasets. The NuScenes dataset contains 1000 driving sequences. Each sequence has a length of roughly 20 seconds and contains keyframes sampled at 2Hz. 
We follow the official data split setting by training our model with 700 sequences and report the results of the 150 validation sequences. For the KITTI dataset, we follow GNN3DMOT~\cite{weng2020gnn3dmot}'s split setting, which contains 10 training sequences and 11 validation sequences. For all our experiments, we train our modules by using the Adam~\cite{kingma2015adam} optimizer with the initial learning rate 0.001 for 10 epochs.


\subsection{Evaluation Metrics}
To evaluate our algorithm performance, we use the \textit{ Average Multi-Object Tracking Accuracy} (AMOTA), which is also the main evaluation metric used in The NuScenes Tracking Challenge ~\cite{caesar2019nuscenes}.
AMOTA is the average of tracking accuracy at different recall thresholds, defined as follows:
{\small
\begin{equation}
    \mathbf{AMOTA} = \frac{1}{n-1} \sum_{r \in \{ \frac{1}{n-1},  \frac{2}{n-1}, ... , 1\} } \mathbf{MOTAR},
\end{equation}
}
where $n=40$ is the number of sample points. And $r$ is the sampled recall threshold. MOTAR is the Recall-Normalized Multi-Object Tracking Accuracy, defined as the follows:
\begin{equation}
    \resizebox{.85\hsize}{!}{$\mathbf{MOTAR} = \mathbf{max}(0, 1 -  \frac{IDS_r + FP_r + FN_r - (1-r) \times P}{r \times P}),$}
\end{equation}

where $P$ is the number of ground-truth positives, $IDS_r$ is the number of identity switches, $FP_r$ is the number of false positives, and $FN_r$ is the number of false negatives.

For KITTI~\cite{geiger2012kitti}, we also report the standard \textit{Multi-Object Tracking Accuracy} (MOTA), defined as the follows: 
\begin{equation}
    \resizebox{.5\hsize}{!}{$\mathbf{MOTA} = 1 -  \frac{IDS + FP + FN}{P},$}
\end{equation}
where $IDS$, $FP$, and $FN$ are the numbers of identity switches, false positives, and false negatives sampled at a single best recall threshold.

\subsection{Quantitative Results}


We report our results on the NuScenes validation set in Table~\ref{tab:val_evaluation_results}. Our proposed tracking method uses CenterPoint~\cite{yin2020center}'s 3D object detection results at each frame as the input to our Kalman Filters.
For a fair comparison with state-of-the-art methods~\cite{yin2020center,chiu2020probabilistic, weng2019ab3dmot}, we also include our tracking method's quantitative results when using only LiDAR as input. From Table~\ref{tab:val_evaluation_results}, we can see that the quality of input detections is critical to the final tracking performance. CenterPoint~\cite{yin2020center} provides better 3D object detection results than  MEGVII~\cite{zhu2019megvii} on the NuScenes Detection Challenge~\cite{caesar2019nuscenes}. 



As can be seen in the last two rows in Table \ref{tab:val_evaluation_results}, when using exactly the same 3D only LiDAR input, our tracking method outperforms CenterPoint~\cite{yin2020center} and ProbabilisticTracking~\cite{chiu2020probabilistic}. We conclude that our model is able to use the 3D LiDAR point cloud data to learn the fine-grained geometric features, and that our model also successfully learns the effective combination weightings of the geometric feature distance and the Mahalanobis distance. Moreover, by fusing features from both LiDAR and image data, our method can further improve the overall AMOTA and results in a 2.8 performance gain compared with the previous state-of-the-art  CenterPoint~\cite{yin2020center}. This performance gain shows that our model is able to learn how to effectively fuse the 3D LiDAR point cloud and 2D camera image input together to achieve better overall tracking accuracy. However, our model does not achieve significant improvement for pedestrians. That could be due to the large change that each pedestrian's appearance and geometric features go through over time as they change pose. Other objects undergo zero or no deformation.

We also compare our model with other multi-modal tracking models: GNN3DMOT~\cite{weng2020gnn3dmot} and PnPNet~\cite{ liang2020pnpnet} in Table~\ref{tab:val_evaluation_partial} for  NuScenes~\cite{caesar2019nuscenes} and Table~\ref{tab:val_evaluation_kitti} for KITTI~\cite{geiger2012kitti}.


\subsection{Ablation Study}
We provide an ablation analysis of the different trainable modules to better understand their contribution to the overall system performance: the \textbf{Distance Combination Module}, the \textbf{Track Initialization Module} and the \textbf{Feature Fusion Module}. We report our results in Table~\ref{tab:val_ablation_results}. 
We note that both the \textbf{Distance Combination Module} and the \textbf{Track Initialization Module} yield consistent improvements over the baseline, with the highest numbers achieved when both modules are enabled. Additionally, we record a consistent increase in performance when fusing 2D and 3D features, allowing us to conclude that our model successfully learns how to leverage both appearance and geometry features.

\begin{table*}[t!]
\small
\caption{Ablation results for the validation set of NuScenes~\cite{caesar2019nuscenes}: evaluation in terms of overall AMOTA and individual AMOTA for each object category, in comparison with variations of our proposed method. All variations use CenterPoint~\cite{yin2020center}'s object detection results as input. In each column, the best-obtained results are typeset in boldface.)
\vspace{-10pt}}
\label{tab:val_ablation_results}
\begin{center}
\begin{tabular}{ l|l|l|cccccccc}
  \hline
  Tracking method & Modalities & Overall & bicycle & bus & car & motorcycle & pedestrian & trailer & truck \\
  \hline
  \hline
  Distance Combination Module only & 3D & 67.1 & 46.3 & 81.9 & 84.2 & 63.8 & 74.9 & \textbf{53.5} & \textbf{65.4} \\
  Track Initialization Module only & 3D & 66.2 & 45.1 & 78.4 & 84.2 & 66.6 & 75.1 & 52.7 & 61.2 \\
  Our proposed method & 3D & 67.7 & 47.0 & 81.9 & 84.2 & 66.8 & 75.2 & \textbf{53.5} & \textbf{65.4} \\
  \hline
  Distance Combination Module only & 2D + 3D & 67.6 & 46.5 & \textbf{82.0} & \textbf{84.3} & 65.4 & 76.3 & 53.1 & \textbf{65.4} \\
  Track Initialization Module only & 2D + 3D & 67.4 & 48.6 & 80.4 & 81.6 & 68.4 & 75.3 & 53.3 & 64.5 \\
  Our proposed method & 2D + 3D & \textbf{68.7} & \textbf{49.0} & \textbf{82.0} & \textbf{84.3} & \textbf{70.2} & \textbf{76.6} & 53.4 & \textbf{65.4} \\
  \hline
\end{tabular}
\end{center}
\vspace{-10pt}
\end{table*}

\subsection{Qualitative Results}
\begin{figure*}[h!]
\vspace{-5pt}
        \centering
        \begin{subfigure}[b]{0.23\textwidth}
            \centering 
            \includegraphics[width=\textwidth]{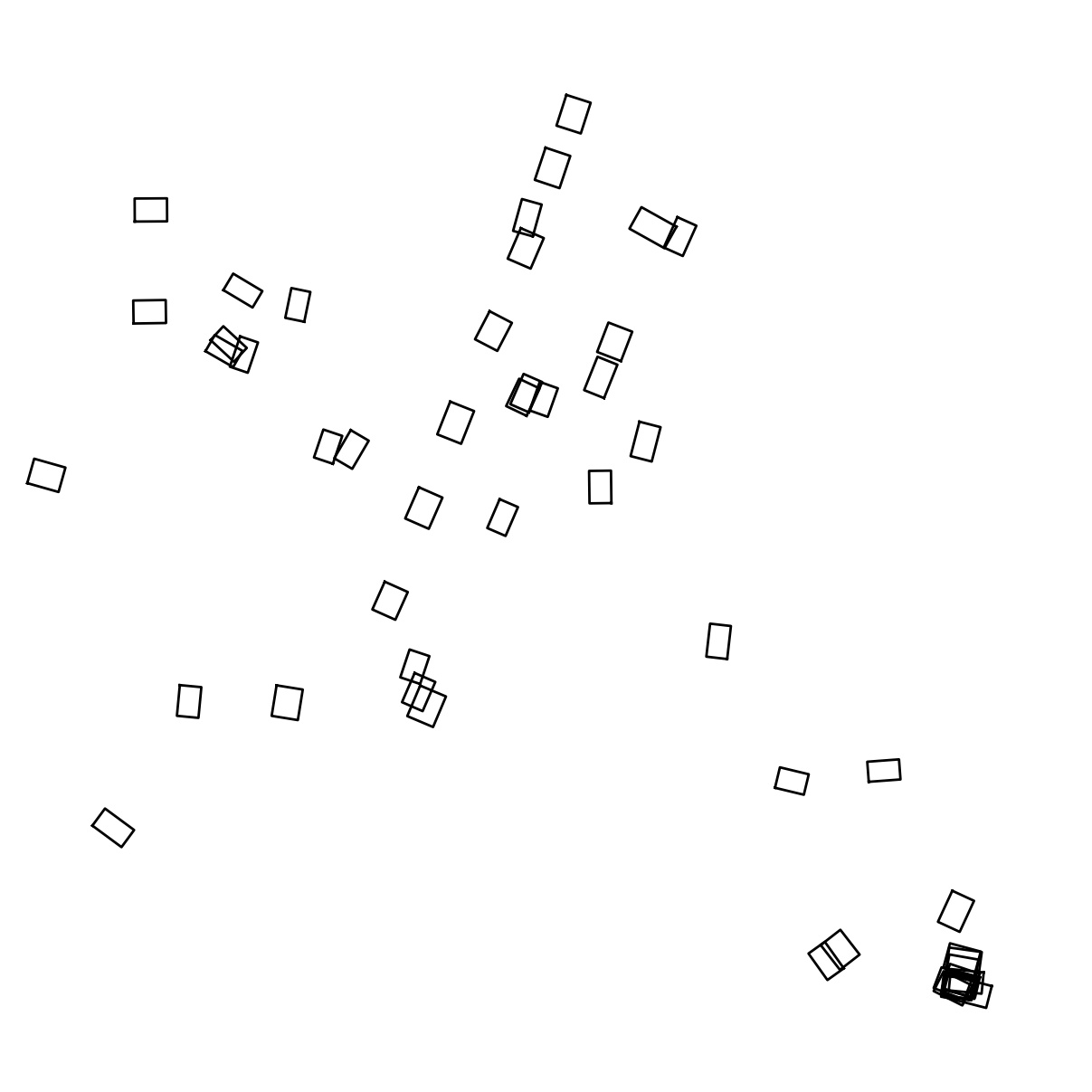}
            \caption[]%
            {{Input detection}}    
            \label{fig:detection_201_motorcycle}
        \end{subfigure}
        \hfill
        \begin{subfigure}[b]{0.23\textwidth}  
            \centering 
            \includegraphics[width=\textwidth]{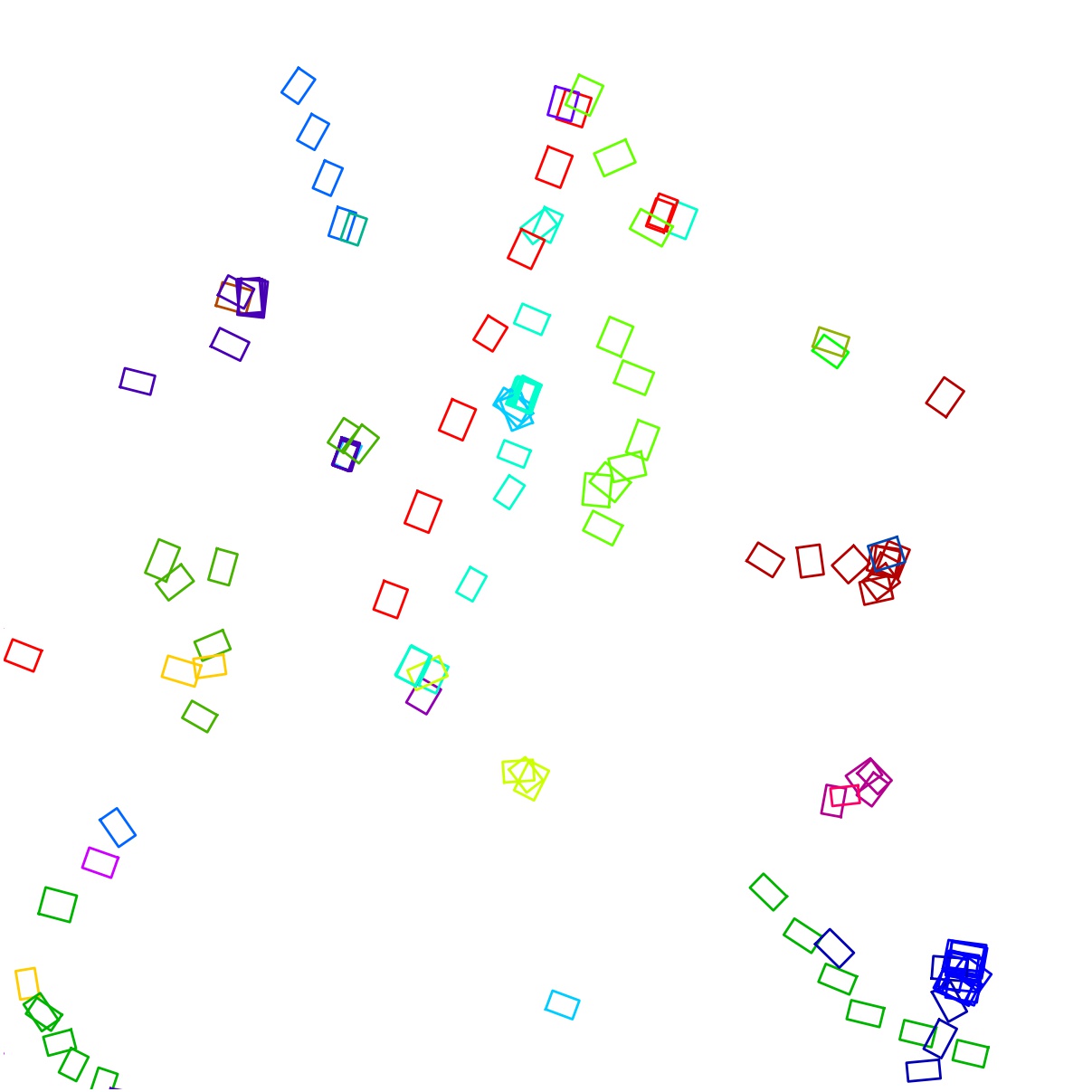}
            \caption[]%
            {{ CenterPoint~\cite{yin2020center}}}    
            \label{fig:center_point_201_motorcycle}
        \end{subfigure}
        \hfill
        \begin{subfigure}[b]{0.23\textwidth}
            \centering
            \includegraphics[width=\textwidth]{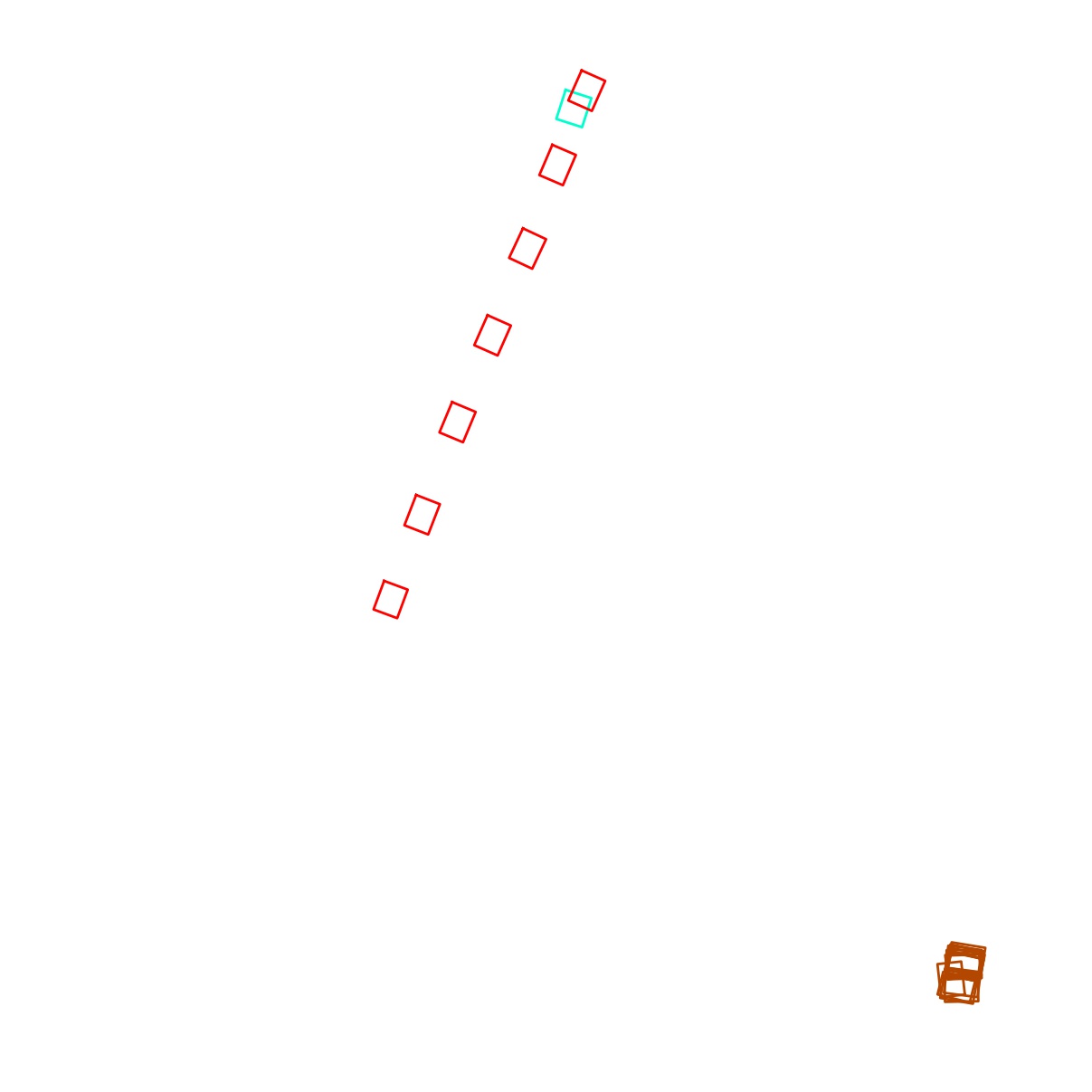}
            \caption[]%
            {{Our proposed method}}    
            \label{fig:ours_201_motorcycle}
        \end{subfigure}
        \hfill
        \begin{subfigure}[b]{0.23\textwidth}
            \centering 
            \includegraphics[width=\textwidth]{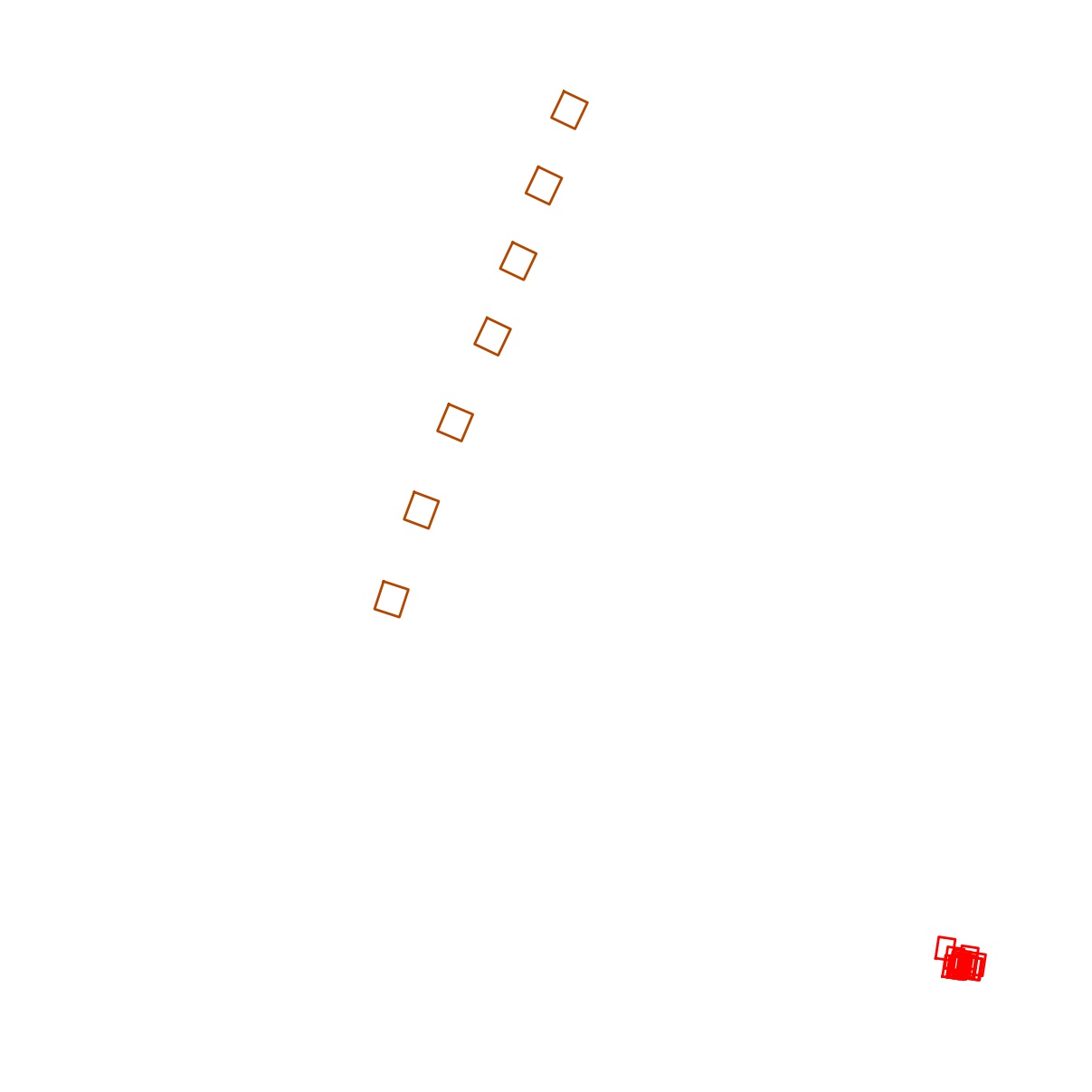}
            \caption[]%
            {{Ground-truth}}    
            \label{fig:gt_201_motorcycle}
        \end{subfigure}
        \hfill
        \caption[]
        {Bird-eye-view tracking visualization of motorcycles. We plot the bounding boxes from every frame of the same driving sequence in each sub-figure. Different colors represent different tracking ids in the tracking results, and indicate different instances of objects in the ground-truth annotation. (a): input detection bounding boxes provided by CenterPoint~\cite{yin2020center}'s object detector. (b): CenterPoint~\cite{yin2020center}'s tracking result. (c): our proposed method's tracking result. (d): ground-truth annotation.  
        Our tracking result has significantly fewer false-positive bounding boxes compared with CenterPoint~\cite{yin2020center}'s result. Our tracking result is also closer to the ground-truth annotation.
        } 
        \label{fig:visual_motorcycle}
    \end{figure*}

\begin{figure*}[h!]
        \centering
        \begin{subfigure}[b]{0.24\textwidth}
            \centering 
            \includegraphics[width=\textwidth]{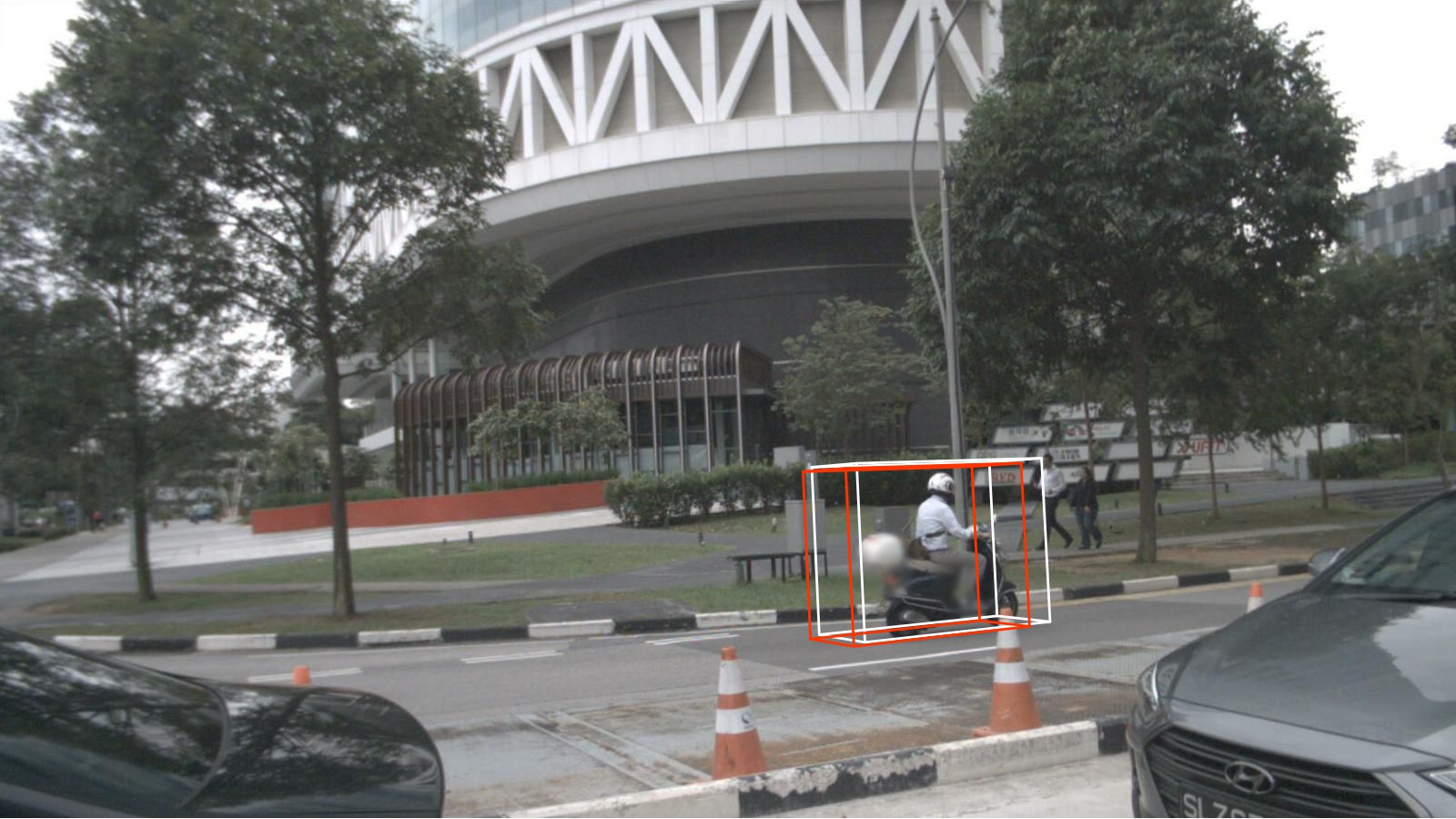}
            \caption[]%
            {{Sequence 0, Frame 1}}    
            \label{fig:visual_sequence_0_frame_1}
        \end{subfigure}
        \hfill
        \begin{subfigure}[b]{0.24\textwidth}  
            \centering 
            \includegraphics[width=\textwidth]{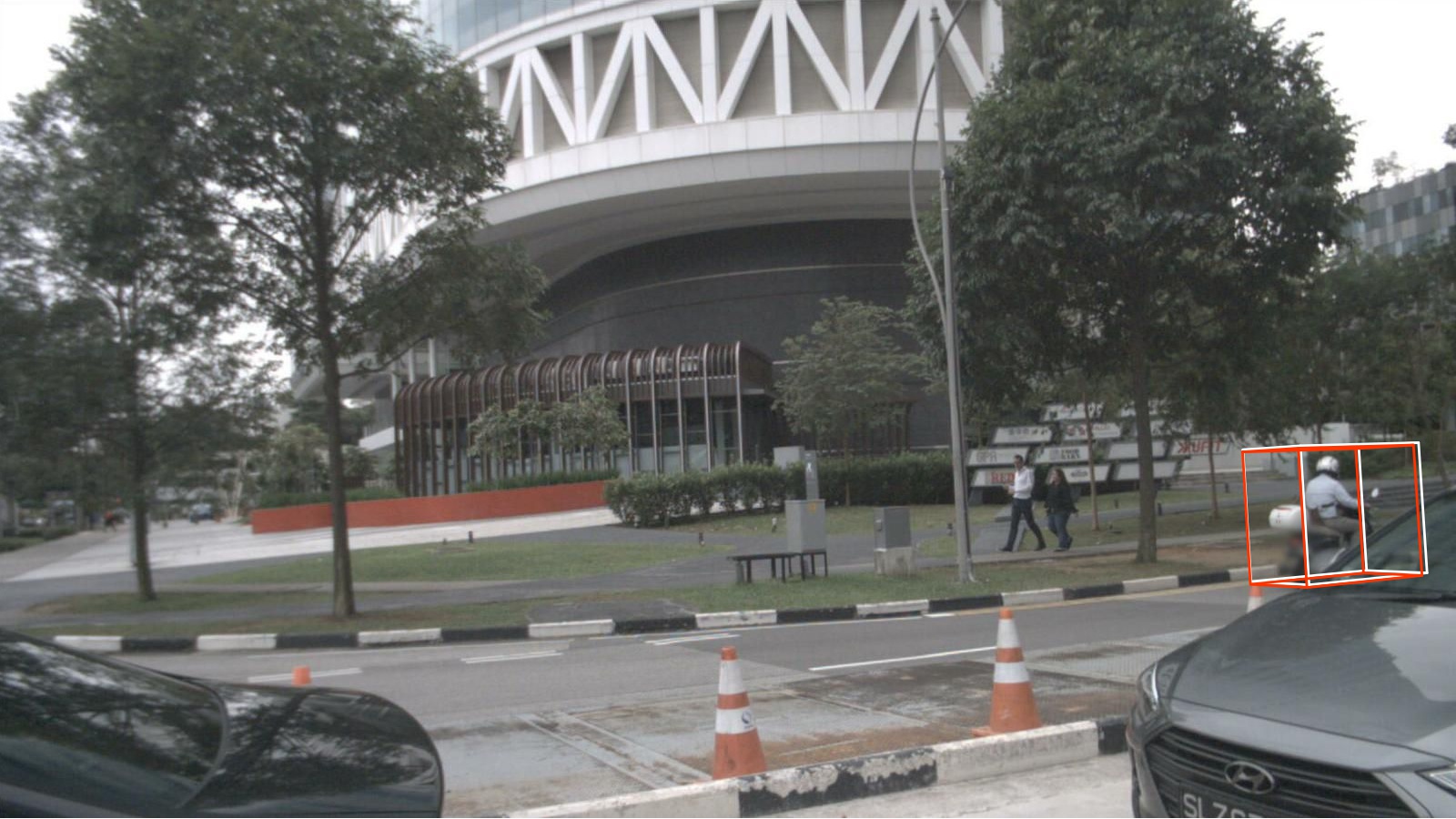}
            \caption[]%
            {{Sequence 0, Frame 2}}    
            \label{fig:visual_sequence_0_frame_2}
        \end{subfigure}
        \hfill
        \begin{subfigure}[b]{0.24\textwidth}
            \centering 
            \includegraphics[width=\textwidth]{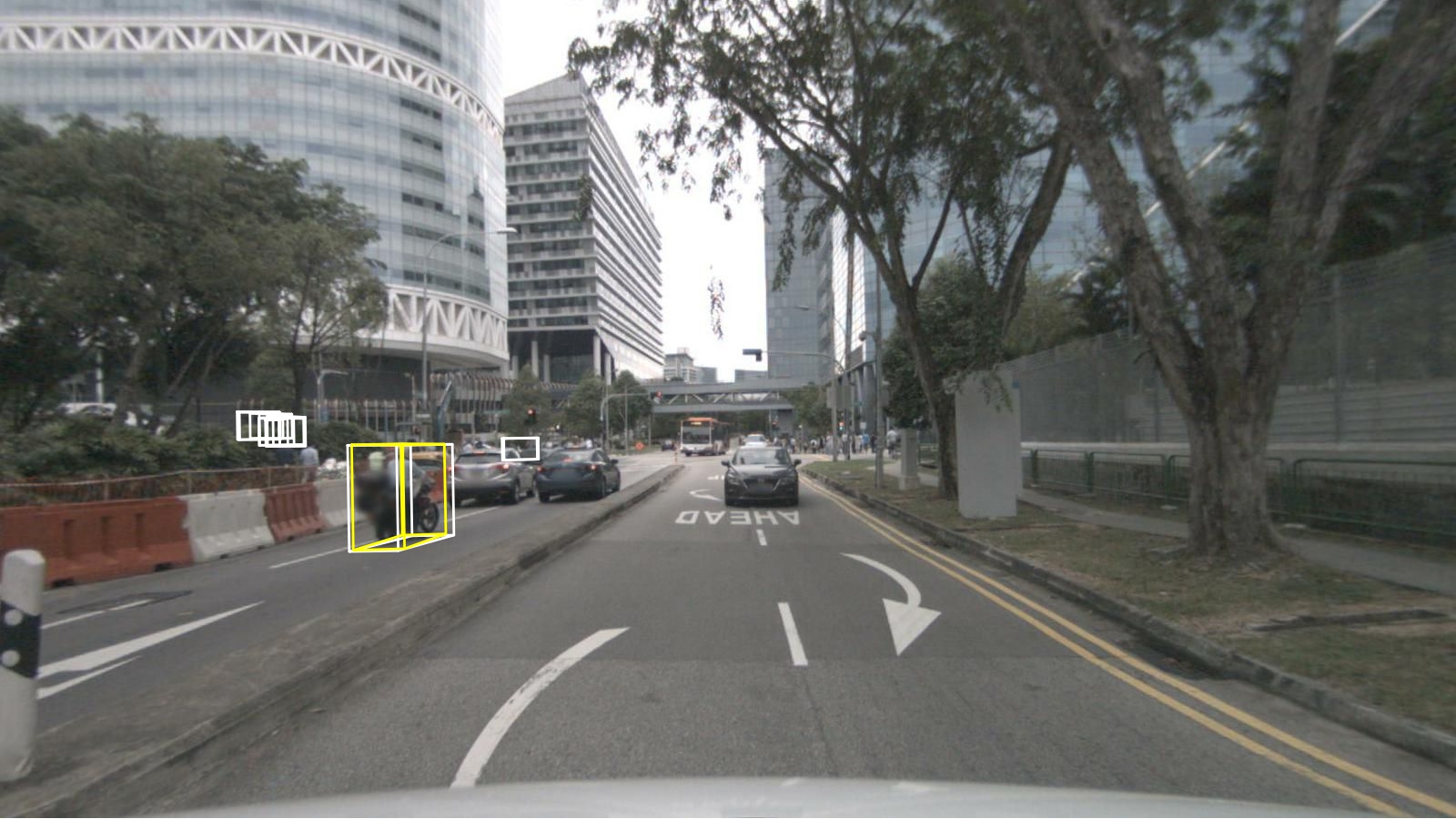}
            \caption[]%
            {{Sequence 1, Frame 28}}    
            \label{fig:visual_sequence_1_frame_28}
        \end{subfigure}
        \hfill
        \begin{subfigure}[b]{0.24\textwidth}  
            \centering 
            \includegraphics[width=\textwidth]{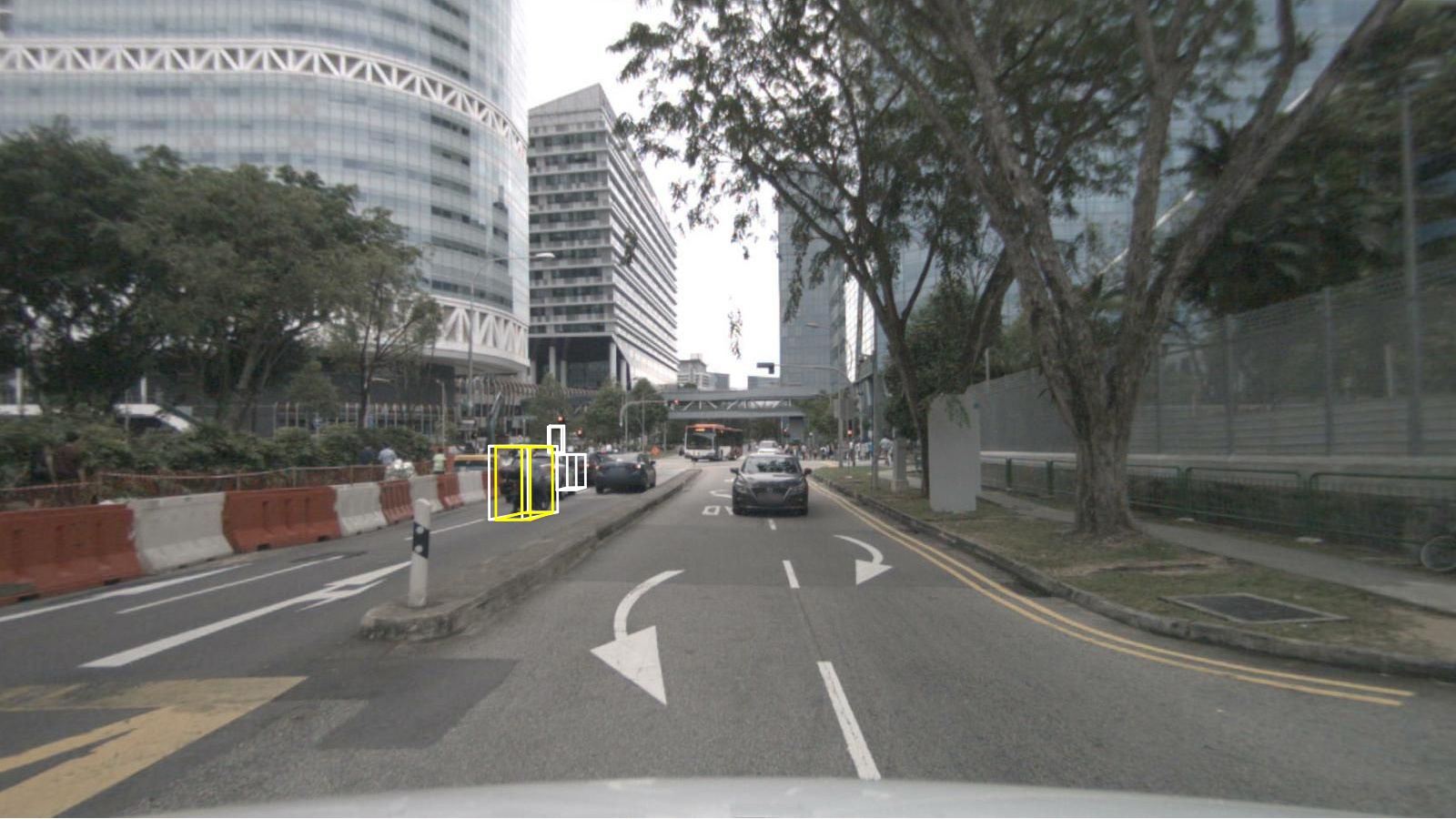}
            \caption[]%
            {{Sequence 1, Frame 29}}    
            \label{fig:visual_sequence_1_frame_29}
        \end{subfigure}
        \hfill
        \caption[]
        {Tracking visualization of motorcycles projected to camera images. (a), (b) are two consecutive frames in Sequence 0. (c), (d) are from Sequence 1. The color boxes are the tracking results. Different colors indicate different tracking ids. The white boxes represent the detections. 
        Our model can accurately track the motorcycles in the red bounding boxes in Sequence $0$ and the yellow bounding boxes in Sequence $1$. 
        In Sequence $0$, our \textbf{Distance Combination Module} learns to generate a larger positive $\alpha = 2.594$ value, because the appearance features seem to provide strong information to match the detected motorcycles across these consecutive frames. In Sequence $1$, our model generates a smaller $\alpha = 1.802$, potentially because the bounding boxes are smaller and the image is more blurred.
        Our \textbf{Track Initialization Module} also correctly decides not to initialize new tracks for the false-positive detections in (c) Sequence $1$ Frame $28$ .  
        \vspace{-15pt}} 
        \label{fig:visual_rgb}
    \end{figure*}



As indicated in Table~\ref{tab:val_evaluation_results}, we note a significant improvement on specific classes (i.e. more than 10 on the motorcycle class). In Figure~\ref{fig:visual_motorcycle}, we plot the bounding boxes of motorcycles from every frame of the same driving sequence on BEV images, with different colors representing different tracking ids, and compare against~\cite{yin2020center}.  From Figure~\ref{fig:visual_motorcycle}, we see that our tracking results have significantly fewer false-positive bounding boxes compared with~\cite{yin2020center}. CenterPoint~\cite{yin2020center} relies on the center Euclidean distance, and any unmatched detection box is always initialized as a new track. Conversely, our \textbf{Track Initialization Module} is designed to decide whether to initialize a new track based on the fusion of the 3D LiDAR and 2D image features. Additionally, our method uses the Kalman Filter to refine the bounding box locations, orientations, and scales based on the past observation, while~\cite{yin2020center} directly uses the potentially noisy detection boxes as the tracking results, without utilizing the past observations. 

While quantitatively we record an 11.0 increase in AMOTA on the motorcycle class compared to CenterPoint~\cite{yin2020center}, qualitatively, this translates to a significant reduction in the number of false positive tracks which are not penalized too much by the AMOTA metric, but which can be crucial for decision making. The main reason behind this discrepancy between qualitative and quantitative results is that most of the false-positive tracks are composed of false-positive detection boxes with low confidence scores. AMOTA starts to sample tracks from those with higher confidence scores. Therefore, a large number of false-positive tracks with low confidence scores will not affect AMOTA too much (for details on AMOTA please refer to~\cite{caesar2019nuscenes}).

Figure~\ref{fig:visual_rgb} visualizes our results of motorcycles projected to camera images. (a), (b) are two consecutive frames in Sequence $0$. (c), (d) are from Sequence $1$. The white boxes represent the detections. The colored boxes indicate the tracking results with color-coded tracking ids. Our model accurately tracks the motorcycles in both sequences.
In Sequence $0$, our \textbf{Distance Combination Module} predicts a larger positive $\alpha = 2.594$ value for the tracked motorcycle, indicating a more reliable feature distance. This is expected as the corresponding objects are large and clearly captured in the 2D images. In Sequence $1$ where the objects are small and blurred, the module predicts a smaller $\alpha = 1.802$.
Additionally, our \textbf{Track Initialization Module} also correctly decides not to initialize new tracks for the false-positive detections in Sequence $1$ Frame $28$.


\section{Conclusion}
In this paper, we propose an online probabilistic, multi-modal, multi-object tracking algorithm for autonomous driving. Our model learns to fuse 2D camera image and 3D LiDAR point cloud features. And these fused features are then used to learn the effective weightings for combining the deep feature distance with the Mahalanobis distance for better data association. Our model also learns to manage track life cycles in a data-driven approach. We evaluate our proposed method on the NuScenes~\cite{caesar2019nuscenes} and KITTI~\cite{geiger2012kitti} datasets. Our method outperforms the state-of-the-art baselines that use the same object detector both quantitatively and qualitatively.

For future work, we are looking to include additional modalities (e.g. map data), as well as novel object detectors. There is also the potential for learning better motion models per category that could further improve data association. And finally, we may leverage differentiable filtering frameworks to fine-tune motion and observation models end-to-end. 


\clearpage
\balance
{\small
\bibliographystyle{IEEEtran}
\bibliography{egbib}

\begin{thebibliography}{10}
\providecommand{\url}[1]{#1}
\csname url@rmstyle\endcsname
\providecommand{\newblock}{\relax}
\providecommand{\bibinfo}[2]{#2}
\providecommand\BIBentrySTDinterwordspacing{\spaceskip=0pt\relax}
\providecommand\BIBentryALTinterwordstretchfactor{4}
\providecommand\BIBentryALTinterwordspacing{\spaceskip=\fontdimen2\font plus
\BIBentryALTinterwordstretchfactor\fontdimen3\font minus
  \fontdimen4\font\relax}
\providecommand\BIBforeignlanguage[2]{{%
\expandafter\ifx\csname l@#1\endcsname\relax
\typeout{** WARNING: IEEEtran.bst: No hyphenation pattern has been}%
\typeout{** loaded for the language `#1'. Using the pattern for}%
\typeout{** the default language instead.}%
\else
\language=\csname l@#1\endcsname
\fi
#2}}

\bibitem{yin2020center}
T.~Yin, X.~Zhou, and P.~Kr{\"a}henb{\"u}hl, ``Center-based 3d object detection
  and tracking,'' in \emph{CVPR}, 2021.

\bibitem{chiu2020probabilistic}
H.-k. Chiu, A.~Prioletti, J.~Li, and J.~Bohg, ``Probabilistic 3d multi-object
  tracking for autonomous driving,'' \emph{arXiv preprint arXiv:2001.05673},
  2020.

\bibitem{mahalanobis1936distance}
P.~C. Mahalanobis, ``On the generalized distance in statistics,''
  \emph{Proceedings of the National Institute of Sciences of India}, 1936.

\bibitem{liang2020pnpnet}
M.~Liang, B.~Yang, W.~Zeng, Y.~Chen, R.~Hu, S.~Casas, and R.~Urtasun, ``Pnpnet:
  End-to-end perception and prediction with tracking in the loop,'' in
  \emph{CVPR}, 2020.

\bibitem{weng2020gnn3dmot}
X.~Weng, Y.~Wang, Y.~Man, and K.~Kitani, ``Gnn3dmot: Graph neural network for
  3d multi-object tracking with multi-feature learning,'' in \emph{CVPR}, 2020.

\bibitem{kalman1960filter}
R.~E. Kalman, ``A new approach to linear filtering and prediction problems,''
  \emph{Journal of Basic Engineering}, 1960.

\bibitem{weng2019ab3dmot}
X.~Weng, J.~Wang, D.~Held, and K.~Kitani, ``{3D Multi-Object Tracking: A
  Baseline and New Evaluation Metrics},'' \emph{IROS}, 2020.

\bibitem{shenoi2020jrmot}
A.~Shenoi, M.~Patel, J.~Gwak, P.~Goebel, A.~Sadeghian, H.~Rezatofighi,
  R.~Martin-Martin, and S.~Savarese, ``Jrmot: A real-time 3d multi-object
  tracker and a new large-scale dataset,'' \emph{IROS}, 2020.

\bibitem{caesar2019nuscenes}
H.~Caesar, V.~Bankiti, A.~H. Lang, S.~Vora, V.~E. Liong, Q.~Xu, A.~Krishnan,
  Y.~Pan, G.~Baldan, and O.~Beijbom, ``nuscenes: A multimodal dataset for
  autonomous driving,'' \emph{CVPR}, 2020.

\bibitem{geiger2012kitti}
A.~Geiger, P.~Lenz, and R.~Urtasun, ``Are we ready for autonomous driving? the
  kitti vision benchmark suite,'' in \emph{CVPR}, 2012.

\bibitem{shi2019pointrcnn}
S.~Shi, X.~Wang, and H.~Li, ``Pointrcnn: 3d object proposal generation and
  detection from point cloud,'' in \emph{CVPR}, 2019.

\bibitem{zhou2020tracking}
X.~Zhou, V.~Koltun, and P.~Kr{\"a}henb{\"u}hl, ``Tracking objects as points,''
  \emph{ECCV}, 2020.

\bibitem{chen2016monocular}
X.~Chen, K.~Kundu, Z.~Zhang, H.~Ma, S.~Fidler, and R.~Urtasun, ``Monocular 3d
  object detection for autonomous driving,'' in \emph{CVPR}, 2016.

\bibitem{brazil2019m3d}
G.~Brazil and X.~Liu, ``M3d-rpn: Monocular 3d region proposal network for
  object detection,'' in \emph{ICCV}, 2019.

\bibitem{zhu2019megvii}
B.~{Zhu}, Z.~{Jiang}, X.~{Zhou}, Z.~{Li}, and G.~{Yu}, ``{Class-balanced
  Grouping and Sampling for Point Cloud 3D Object Detection},'' \emph{arXiv
  preprint arXiv:1908.09492}, 2019.

\bibitem{zhou2018voxelnet}
Y.~Zhou and O.~Tuzel, ``Voxelnet: End-to-end learning for point cloud based 3d
  object detection,'' in \emph{CVPR}, 2018.

\bibitem{yan2018second}
Y.~Yan, Y.~Mao, and B.~Li, ``Second: Sparsely embedded convolutional
  detection,'' in \emph{Sensors}, 2018.

\bibitem{yan2018pixor}
B.~Yang, W.~Luo, and R.~Urtasun, ``Pixor: Real-time 3d object detection from
  point clouds,'' in \emph{CVPR}, 2018.

\bibitem{lang2019pointpillar}
A.~H. Lang, S.~Vora, H.~Caesar, L.~Zhou, J.~Yang, and O.~Beijbom,
  ``Pointpillars: Fast encoders for object detection from point clouds,'' in
  \emph{CVPR}, 2019.

\bibitem{liang2019multi}
M.~Liang, , B.~Yang, Y.~Chen, R.~Hu, and R.~Urtasun, ``Multi-task multi-sensor
  fusion for 3d object detection,'' in \emph{CVPR}, 2019.

\bibitem{vora2020pointpainting}
S.~Vora, A.~H. Lang, B.~Helou, and O.~Beijbom, ``Pointpainting: Sequential
  fusion for 3d object detection,'' in \emph{CVPR}, 2020.

\bibitem{qi2018frustum}
C.~R. Qi, W.~Liu, C.~We, H.~Su, and L.~J. Guibas, ``Frustum pointnets for 3d
  object detection from rgb-d data,'' in \emph{CVPR}, 2018.

\bibitem{hu2019joint}
H.-N. Hu, Q.-Z. Cai, D.~Wang, J.~Lin, M.~Sun, P.~Krähenbühl, T.~Darrell, and
  F.~Yu, ``Joint monocular 3d detection and tracking,'' in \emph{ICCV}, 2019.

\bibitem{qi2019pointnet}
C.~R. Qi, H.~Su, K.~Mo, and L.~J. Guibas, ``Pointnet: Deep learning on point
  sets for 3d classification and segmentation,'' in \emph{CVPR}, 2017.

\bibitem{qi2017pointnetplusplus}
C.~R. Qi, L.~Yi, H.~Su, and L.~J. Guibas, ``Pointnet++: Deep hierarchical
  feature learning on point sets in a metric space,'' \emph{NeurIPS}, 2017.

\bibitem{bewley2016simple}
A.~Bewley, Z.~Ge, L.~Ott, F.~Ramos, and B.~Upcroft, ``Simple online and
  realtime tracking,'' in \emph{ICIP}, 2016.

\bibitem{lin2014microsoft}
T.-Y. Lin, M.~Maire, S.~Belongie, L.~Bourdev, R.~Girshick, J.~Hays, P.~Perona,
  D.~Ramanan, C.~L. Zitnick, and P.~Dollár, ``Microsoft coco: Common objects
  in context,'' in \emph{ECCV}, 2014.

\bibitem{he2017mask}
K.~He, G.~Gkioxari, P.~Dollár, and R.~Girshick, ``Mask r-cnn,'' in
  \emph{ICCV}, 2017.

\bibitem{kingma2015adam}
D.~Kingma and J.~Ba, ``Adam: A method for stochastic optimization,'' in
  \emph{ICLR}, 2015.

\end{thebibliography}
}

\end{document}